\documentclass[review]{elsarticle}
\usepackage{times}
\usepackage{latexsym}
\usepackage{url}
\usepackage{etoolbox}
\usepackage{graphicx}
\usepackage{url}
\usepackage{amsmath}
\usepackage{amssymb}
\usepackage{color}         
\usepackage{multirow}
\usepackage{comment}
\usepackage{pdflscape}
\usepackage{bibentry}
\usepackage{subcaption}
\usepackage[pagewise]{lineno}
\usepackage{microtype}
\usepackage{graphicx}
\usepackage{amssymb}
\usepackage{amsmath}
\usepackage{multirow}
\usepackage{graphicx}
\usepackage{multirow}
\usepackage{amssymb}
\usepackage{amsmath}
\usepackage{algpseudocode} 
\usepackage[ruled,linesnumbered]{algorithm2e}

\journal{Expert Systems with Applications}

\author{Meher Bhardwaj$^1$, Hrishikesh Ethari$^1$, and Dennis Singh Moirangthem*$^2$\\
Email: meher@iiitmanipur.ac.in, hrishikesh@iiitmanipur.ac.in, and dennis@nitmanipur.ac.in*}

\address{$^1$Department of Computer Science and Engineering\\ Indian Institute of Information Technology Senapati, Manipur\\
Imphal, Manipur, India - 795002}

\address{$^2$Department of Computer Science and Engineering\\ National Institute of Technology Manipur\\
Imphal, Manipur, India - 795004}

\begin{document}
\begin{frontmatter}    

\title{EzSQL: An SQL intermediate representation for improving SQL-to-text Generation}

\begin{abstract}
The SQL-to-text generation task traditionally uses template base, Seq2Seq, tree-to-sequence, and graph-to-sequence models. Recent models take advantage of pre-trained generative language models for this task in the Seq2Seq framework. However, treating SQL as a sequence of inputs to the pre-trained models is not optimal. 
In this work, we put forward a new SQL intermediate representation called EzSQL to align SQL with the natural language text sequence. EzSQL simplifies SQL queries and brings them closer to natural language text by modifying operators and keywords, which can usually be described in natural language. EzSQL also removes the need for set operators. Our proposed SQL-to-text generation model uses EzSQL as the input to a pre-trained generative language model for generating the text descriptions. We demonstrate that our model is an effective state-of-the-art method to generate text narrations from SQL queries on the WikiSQL and Spider datasets. We also show that by generating pretraining data using our SQL-to-text generation model, we can enhance the performance of Text-to-SQL parsers.

\end{abstract}


\end{frontmatter}
*Corresponding author\\

\section{Introduction}
An SQL-to-text model automatically generates human-like text descriptions by interpreting the meaning signified by the structured query language (SQL) query. This task has been crucial for natural language to interact with a database, as it helps users who are unfamiliar with SQL to understand the queries. A newly found application of this task is training data generation for enhancing the performance of Text-to-SQL parsers \cite{shi2021learning, wang2021learning}. Earlier attempts were made for the SQL-to-text task in the form of rule-based and template-based \cite{koutrika2010explaining,ngonga2013sorry} models. Further, newer approaches use sequence-to-sequence (Seq2Seq) networks to model the SQL queries and natural language in an interconnected manner \cite{iyer-etal-2016-summarizing}.
Furthermore, several models were also developed based on graph encoding techniques \cite{kipf2016semi,song-etal-2018-graph}, since SQL is designed to express the query intent in a graph-structured manner.

Recently, self-supervised deep contextual language models \cite{devlin-etal-2019-bert, liu2019roberta,lewis-etal-2020-bart} have shown their effective modeling ability for text, demonstrating state-of-the-art results in a series of NLP tasks. These tasks include sequence understanding and language generation.
Previously, \cite{shi2021learning, wang2021learning} have explored the use of BART \cite{lewis-etal-2020-bart} in SQL-to-Text generation. To generate the text utterances, the authors train their SQL-to-Text model, where the input is the original SQL. The input SQL is directly taken as a sequence by the BART tokenizer since the pre-trained model expects a sequence of text as input. However, directly feeding SQL as a sequence is not optimal. 

\begin{figure}[t]
	\centering
\textit{}	\includegraphics[width=0.6\textwidth]{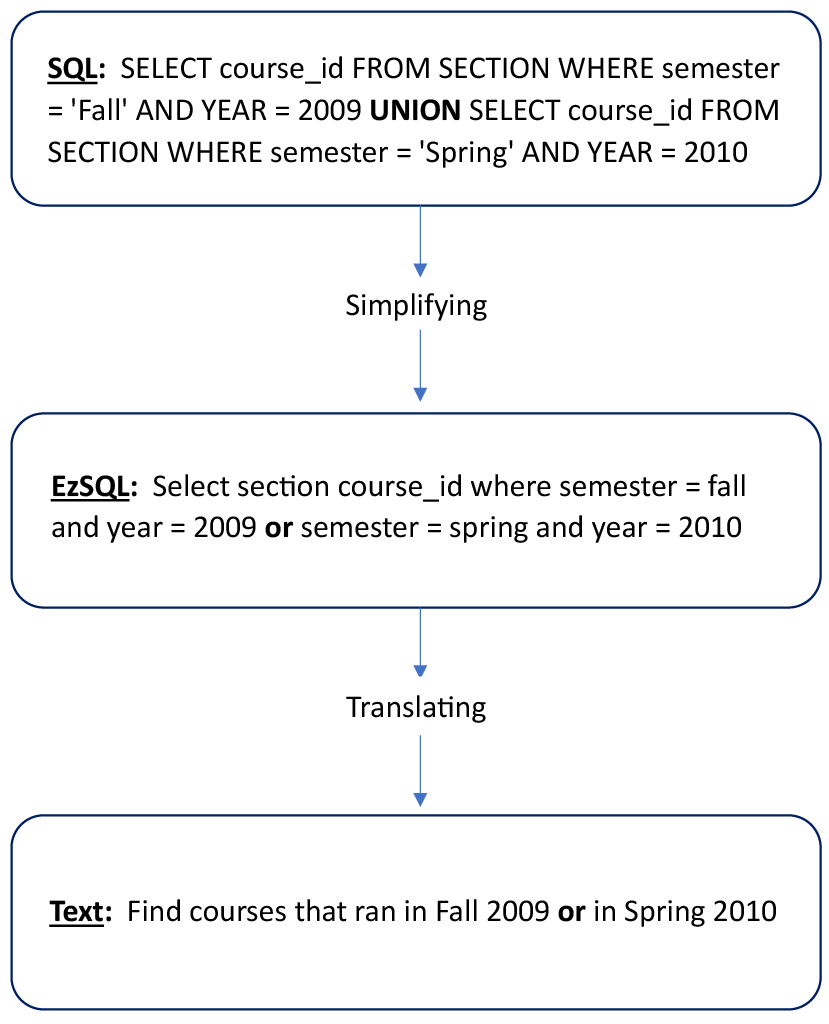}
	\caption{The proposed SQL-to-text model}
	\label{fig:approach}
\end{figure}

In this paper, we present EzSQL, a new intermediate representation of queries focusing on the SQL-to-text task. EzSQL bridges the gap between SQL and natural language (NL). Figure~\ref{fig:approach} presents the EzSQL-based SQL-to-text translation. In the example, EzSQL uses the \textit{OR} conjunctive operator instead of the \textit{UNION} operator, which also removes the multiple \textit{SELECT} clauses, and aligns better with the natural language. These simplification steps in EzSQL make it more suitable as an input to a pre-trained language model for the generation of text descriptions. We propose an SQL-to-Text generator that leverages the capability of a pre-trained language model (BART) to generate text from a given SQL query. We first translate the SQL queries into EzSQL, which is more suitable for the pre-trained language model. The EzSQL is fed as input to BART, where it is directly tokenized by the BART tokenizer without additional pre-processing, and the BART decoder generates the text descriptions.

We evaluate our proposed SQL-to-text model on the WikiSQL and Spider datasets. Experimental results show that our model outperforms the existing model in this task. Furthermore, we also utilize our SQL-to-text model to assist the pre-training data generator in \cite{wang2021learning} for the Text-to-SQL semantic parser. We show that our model was able to enhance the performance of the Text-to-SQL parser pre-trained on data generated using our system.  Our main contributions are summarized as follows:
\begin{itemize}
\item We propose the first SQL intermediate representation for the SQL-to-text task. The representation named EzSQL is simple yet effective in closing the gap between SQL and natural language, enabling the efficient usage of pre-trained language models for text generation.
\item Our proposed model achieves state-of-the-art performance on WikiSQL and Spider datasets in the SQL-to-text setting. We use this model for SQL-Text pair pre-training data generation.
\item We also show that by using the pre-training data generated using our model, we can enhance the performance of Text-to-SQL parsers.
\end{itemize}

\section{Related Work}

\paragraph{SQL-to-text} The rule and template-based SQL-to-text \cite{koutrika2010explaining}, \cite{ngonga2013sorry} models requires human intervention to design templates or rules. Despite the efforts, the generated texts lack the natural feel of the human language. To address this, \cite{iyer-etal-2016-summarizing} proposed a sequence-to-sequence (Seq2Seq) network for this task. However, since SQL is designed to express query intent in a graph-structured manner, \cite{xu2018sql} introduced the Graph-to-Sequence Model for SQL-to-text generation that can capture the graph structure information. Recently, \cite{shi2021learning, wang2021learning} have used large pre-trained language models such as BART~\cite{lewis-etal-2020-bart} for SQL-to-text generation with satisfactory results. 

\paragraph{Intermediate representation} Although IR of SQL for improving the Text-to-SQL has been studied in recent times~\cite{wang-etal-2020-rat,Guo2019,Yu2018-SyntaxSQLNet,shi2021learning,gan2021natural}, IR for SQL-to-text has not been explored previously. In this work, we propose the first IR specifically for the SQL-to-text task.

\paragraph{Data Augmentation} Data augmentation for semantic parsing has attracted a lot of attention in recent years. Earlier attempts include generating new examples using SCFG~\cite{jia-liang-2016-data}, back-translation~\cite{sennrich-etal-2016-improving, dong-etal-2017-learning}, etc. \cite{yu-etal-2018-syntaxsqlnet, yu2020grappa} also use handcrafted SCFG rules to generate new parallel data. In another approach, \cite{zhang2021data} attempted to generate a large number of SQL queries based on an abstract syntax tree grammar and used a hierarchical SQL-to-text generation model in order to synthesize natural language questions. However, this approach uses the traditional seq2seq mechanism for the question generation. Recent approaches, such as~\cite{shi2021learning,wang2021learning} use SQL-to-text for parallel data generation by using powerful conditional text generators such as BART. Following these recent approaches, we also utilize BART for SQL-to-text and data augmentation.

\paragraph{Text-to-SQL} Significant progress has been achieved in text-to-SQL over the past few years. With a suitable amount of in-domain training data, existing models already achieve over 80\% exact matching accuracy \cite{wang2018robust} on single-domain benchmarks like ATIS and GeoQuery. However, annotating NL questions with SQL queries is resource-intensive, which makes it cost-inefficient to collect training examples from all possible databases. \cite{yu-etal-2018-spider} introduced Spider, a cross-database text-to-SQL benchmark evaluates a system using different databases. Recently, several models that target the Spider benchmark have been introduced, including RAT-SQL~\cite{wang-etal-2020-rat} and several other models that build on top of RAT-SQL~\cite{shi2021learning,gan2021natural,wang2021learning}.

\section{Proposed Approach}
We describe our proposed approach in this section.

\subsection{Overview}

Intermediate representation (IR) is widely employed by the Text-to-SQL models to synthesize SQL queries with more complex structures~\cite{wang-etal-2020-rat, Guo2019, Yu2018-SyntaxSQLNet, shi2021learning, gan2021natural}. However, such IRs are either too complicated or have limited coverage of SQL structures due to the restriction that IRs should be converted back to executable SQLs. \\
\begin{figure}[htbp]
	\centering
	\includegraphics[width=1\textwidth]{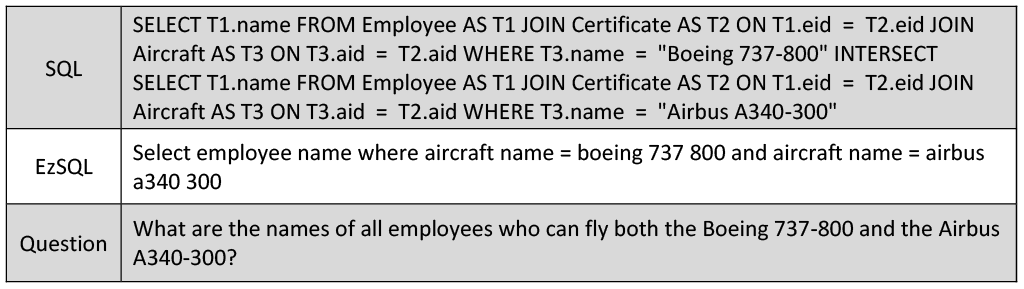}
	\caption{EzSQL simplification exemplified}
	\label{fig:example4}
\end{figure}
Our EzSQL IR focuses on SQL-to-text and hence has no limitations on the simplification process of SQL.
The main goal of EzSQL is to simplify the structure of SQL and bring it as close as possible to natural language while maintaining the context and semantic meaning. Considering the example in Figure~\ref{fig:example4}, the set operator ‘\textit{INTERSECT}’, used to combine \textit{SELECT} statements, is never mentioned in the question. \textit{INTERSECT} is a keyword in SQL that allows the user to seek a combination of the results of multiple functions. Such implementation details are not needed for the BART model to generate the text descriptions, which are rarely mentioned in the target questions~\cite{Guo2019}. 

\begin{figure}[t]
	\centering
	\includegraphics[width=1\textwidth]{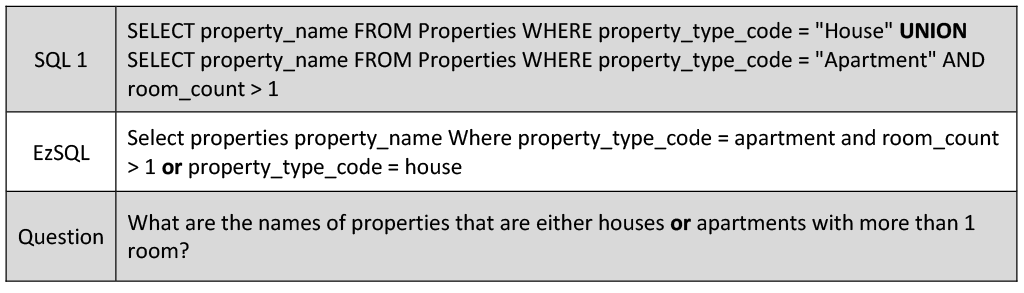}
	\caption{Example demonstrating how EzSQL deals with set operators}
	\label{fig:or_example}
\end{figure}
\begin{figure*}[t]
	\centering
	\includegraphics[width=0.99\textwidth]{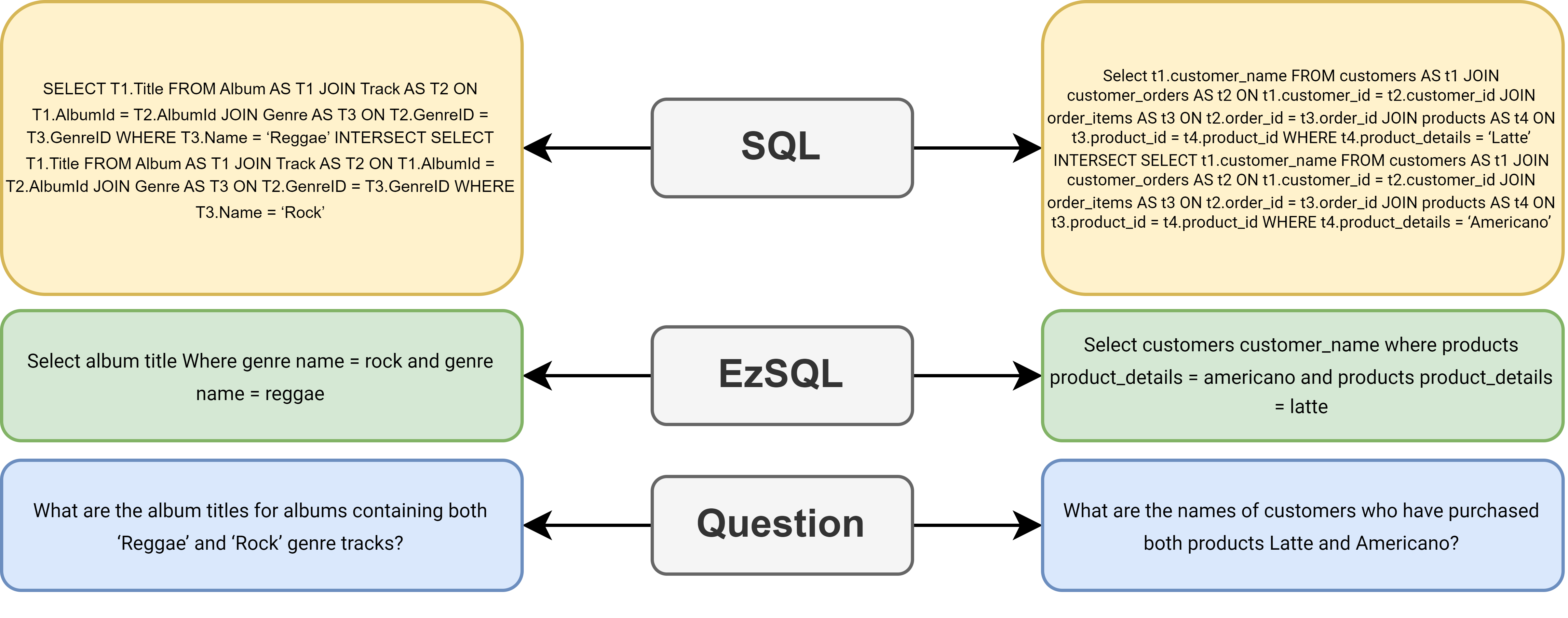}
	\caption{EzSQL brings complex queries significantly closer to the gold text}
	\label{fig:ComplexQueries}
\end{figure*}

Figure~\ref{fig:or_example} compares SQL queries with slight changes in the question. The first question in Figure \ref{fig:or_example} contains an extra condition: `more than 1 room'. This extra condition changes the structure of the entire SQL query. It will not be fair for us to expect end-to-end SQL-to-text models to automatically map \textit{UNION} to \textit{OR} in the text. EzSQL tackles this issue by unifying them into a simple \textit{OR} conjunction.
EzSQL also removes the \textit{JOIN} conditions as shown in Figure~\ref{fig:ComplexQueries}. Therefore, EzSQL improves the representation of SQL for text generation in the following ways: \\
(1) Simplifies the structure of queries with set operators, i.e., \textit{INTERSECT}, \textit{UNION}, and \textit{EXCEPT}. \\ 
(2) Eliminates \textit{JOIN} and nested subqueries. \\
(3) Replaces table and column names with more descriptive names available in the schema.\\
(4) Replaces the keywords \textit{FROM, INTERSECT, UNION, AS} in SQL with appropriate descriptions.\\
(5) Uses DISTINCT and COUNT keywords in a manner that aligns with the natural language. \\
(6) The EzSQL IR reduces the sequence length of long and complex SQL queries. This is crucial in improving the performance and efficiency of the underlying pretrained Transformer models.
\subsection{EzSQL simplification procedure}
\label{subsection:ssql}
\begin{figure}[t!]
	\centering
	\includegraphics[width=0.95\textwidth]{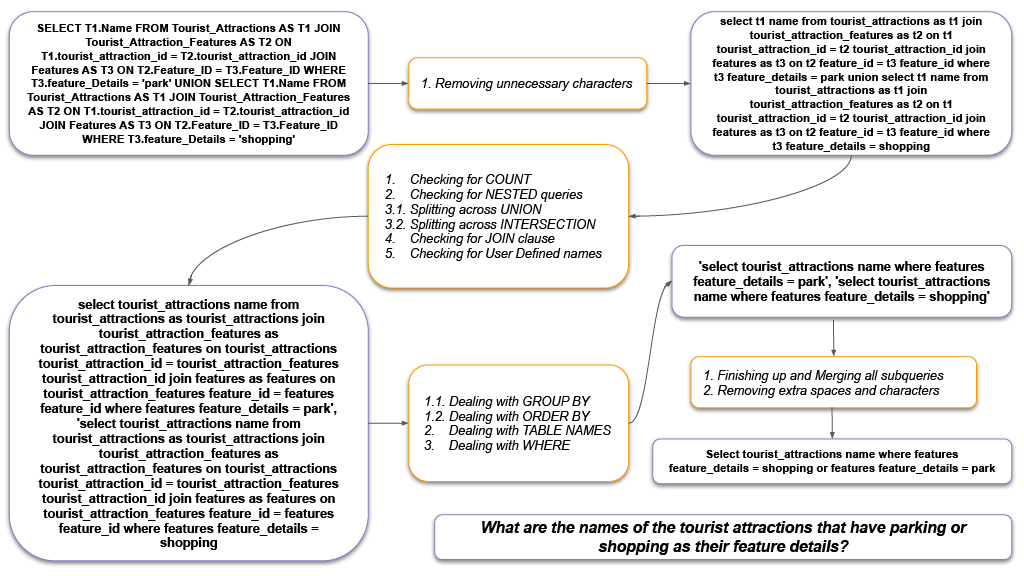}
	\caption{Proposed Approach Example}
	\label{fig:example_99}
\end{figure}

The simplification process in EzSQL can comprise several steps depending on the type and complexity of the query. EzSQL takes several inspirations from existing IR used for Text-to-SQL. The IR in \cite{Yu2018-SyntaxSQLNet} represents SQL statements without \textit{FROM} and \textit{JOIN ON} clauses. \cite{Guo2019} removes the \textit{FROM, JOIN ON} and \textit{GROUP BY} clauses, and combines the \textit{WHERE} and \textit{HAVING} conditions. The recent IR in \cite{gan2021natural} dispenses nested subqueries and set operators. By taking advantage of the existing IRs and combining the generation processes, we implement the EzSQL IR for SQL-to-text with the single goal of representing the SQL statement closest to natural language text. Since our representation does not have any limitations and restrictions for simplification, the implementation is much simpler in the one-way process. The simplification process is represented in figure~\ref{fig:example_99}.

\paragraph{Handling set operators} EzSQL simply concatenates the conditions in a set operator where the conditions can be connected, as shown in Figure~\ref{fig:approach}. However, if the two conditions cannot be connected or they are disjoint, the query is first simplified as given in Figure~\ref{fig:ComplexQueries}. When the \textit{EXCEPT} operator itself forms part of a condition, then the statement is translated into a \textit{WHERE} clause. \newpage

\begin{algorithm}
  \KwData{Query as a string}
  \KwResult{Query with set operators converted into simple queries}

  \While{Query contains `union' keyword}{
    Find the first occurrence of the keyword `union' in the Query\;
    Split the Query into two parts at the position of the keyword\;
    Append the queries in a list called subqueries[]\;
    Set nested\_query\_flag equal to 1   
  }
  \While{Query contains `intersect' keyword}{
    Find the first occurrence of the keyword `intersect' in the Query\;
    Split the Query into two parts at the position of the keyword\;
    Append the queries in a list called subqueries[]\;
    Set nested\_query\_flag equal to 2   
  }
  \caption{Detecting Set Operators}
  \label{algo:1}
\end{algorithm}
\par Algorithm \ref{algo:1} demonstrates how set operators are detected and dealt with for further processing by the EzSQL generation algorithm. When the `Union' operator is detected, the nested\_query\_flag is set to 1, and the query is split across the word `union'.
  In case `Intersect' is detected, the nested\_query\_flag is set to 2, and the query is split across the word `Intersect'. The nested flags shall be used in the later steps so that the semantic meaning is not lost.\\
  These split queries are free of set operators. Since they are independent in nature, it makes them simpler to process, while at the same time inclining them towards natural language. After this, further processing takes place on these split queries independently.
  \newpage
\begin{algorithm}[htbp]
  \KwData{Subqueries as a list of strings}
  \KwResult{Subqueries converted into one query}

  \While{Nested\_query\_flag is equal to 1}{
    Match the elements of the subqueries [] to the first word of difference\;
    Add the word 'or' to the first subquery \;
    Add the words of the remaining subqueries after the first different word onto the first subquery \;
    Set the first subquery as the final string for the next stage  
  }
  \While{Nested\_query\_flag is equal to 2}{
    Match the elements of the subqueries [] to the first word of difference\;
    Add the word 'and' to the first subquery \;
    Add the words of the remaining subqueries after the first different word onto the first subquery \;
    Set the first subquery as the final string for the next stage   
  }
  \caption{Dealing with the removed set operators}
  \label{algo:2}
\end{algorithm}
\par Algorithm \ref{algo:2} demonstrates how the split queries that occurred as the output of algorithm \ref{algo:1} are merged. After all the processing is done, nested\_query\_flag values are checked. A value of 1 signifies the presence of the `Union' operator in the original query, while a value of 2 shows the presence of the `Intersect' operator. \\
The subqueries are thus compared and matched to the first word, which is different in the subqueries. Everything before this word is concatenated, as it is, to the output query. This is usually the point where the `where' clause begins.\\
The portions of the queries occurring after the different words are then concatenated to the output query one by one. After one subquery has been completely concatenated, a word ( \textbf{and} or \textbf{or} ) is inserted, which is based on the value of the nested\_query\_flag. This process is repeated until all subqueries are concatenated to the output query.
\begin{algorithm}
  \KwData{Subqueries as a list of strings}
  \KwResult{Aliases and original names extracted}
  \While{`as' keyword exists in the subquery}{
    Find the occurrence of the keyword `as' in the subqueries\;
    Extract the word occurring before the `as' keyword and append it in the original\_names list\;
    Extract the word occurring after the `as' keyword and append it to the aliases list 
  }
  \caption{Extracting Original names and aliases in the subqueries}
  \label{algo:3}
\end{algorithm}
\par Algorithm \ref{algo:3} sets the foundation for the processes focused on dealing with the aliases. It takes a list of strings as input (called subqueries), and detects the \textbf{`as'} keyword in each of the subqueries. The word occurring before the `as' keyword is the original table name, while the word occurring after it is the alias. Thus, the original table name(s) in each of the subqueries are appended to the original\_names list, while the user-defined names are appended to the aliases list. We take an ordered list for both so that the alias and the corresponding original name occur at the same index in their respective lists.
\begin{algorithm}[htbp]
  \KwData{Subqueries as a list of strings}
  \KwResult{Subqueries with no aliases}

  \While{`as' keyword exists in subquery}{
    Find the occurrence of each user-defined name\;
    Replace each occurrence of the user-defined name with the corresponding original name  
  }
  \caption{Replacing all aliases}
  \label{algo:4}
\end{algorithm}
\par Algorithm \ref{algo:4} makes the use of the lists generated by algorithm \ref{algo:3}. It finds the occurrence of each user-defined name (i.e., alias) and replaces it with the corresponding original table name, by referring to the original\_names list. Thus, all aliases are replaced by their respective original table names.    
\begin{algorithm}[htbp]
  \KwData{Subqueries as a list of strings}
  \KwResult{Subqueries without `as' keyword}

  \While{`as' keyword exists in subquery}{
    Find occurrence of `as' keyword \;
    Delete the `as' keyword and the word occurring after it \;
    Output queries are input for the next stage
  }
  \caption{Removing `as' keyword}
  \label{algo:5}
\end{algorithm}
\par Algorithm \ref{algo:5} is the final step associated with the processing of aliases. The output strings of algorithm \ref{algo:4} are the input for this algorithm. These input strings contain the `as' keyword, and the original table name occurs before and after it ( because of Algorithm \ref{algo:4}). Thus, the `as' keyword is identified, and is removed, along with the word occurring after it. Thus, we obtain queries without aliases and `as' keywords.

\begin{figure}[htbp]
\centering
\includegraphics[width=0.8\textwidth]{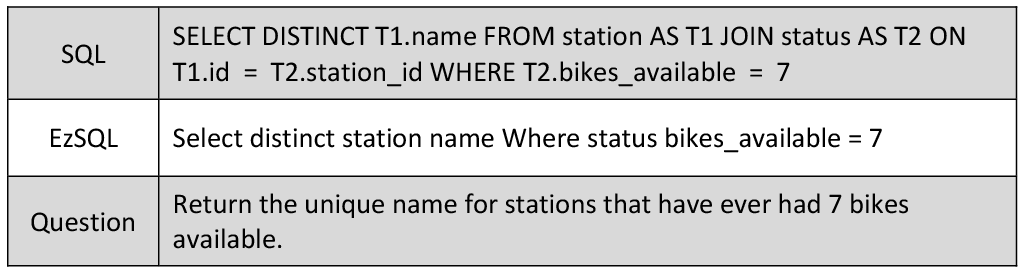}
	\caption{Example demonstrating how EzSQL deals with aliases}
	\label{fig:alias_example}
\end{figure}

\subsection{SQL-to-text Generation Model}\label{section:sql2text}
Our SQL-to-text generation approach takes an SQL query as input and outputs a natural language question.
Our approach is based on a pre-trained sequence-to-sequence BART model~\cite{lewis-etal-2020-bart}. The model consists of an SQL simplifier module to generate EzSQL and a BART encoder-decoder module. The SQL simplifier takes SQL as input and produces the EzSQL IR. The EzSQL IR is then taken by the BART model to generate text descriptions. 

As an IR, the EzSQL need not be a complete text narration and can still contain SQL terms. This flexibility makes the system more robust. We also rely on the text-infilling capability of the pre-trained language model (BART) to generate complete narrations. Template-based systems lack such kind of generation capability.

\section{Experiments}
We first evaluate the performance of our proposed SQL-to-text model on the available benchmarks. We also show the effectiveness of our proposed approach in training data generation for the Text-to-SQL task.

\subsection{SQL-to-text}
We evaluate our model on two datasets, WikiSQL~\cite{zhong2017seq2sql} and Spider~\cite{yu-etal-2018-spider}. WikiSQL consists of a corpus of 87,726 hand-annotated SQL queries and natural language question pairs. These SQL queries are further split into training (61,297 examples), development (9,145 examples), and test sets (17,284 examples). Spider consists of 7000, 1034, and 2147 samples for training, development, and testing, respectively, where 206 databases are split into 146 for training, 20 for development, and 40 for testing. We use the BART-large variant in our model. The BLEU-4 score~\cite{papineni2002bleu} is used as the automatic evaluation metric for this task.
\begin{table}[t!]
\centering
    \resizebox{0.6\columnwidth}{!}{
        \begin{tabular}{l|cc}
            \hline
            Model & \textsc{BLEU}  \\
            \hline
            Template~\cite{xu2018sql} & 15.71  \\
            Seq2Seq~\cite{xu2018sql} & 24.12 \\
            Tree2Seq~\cite{xu2018sql}  & 26.67 \\
            Graph2Seq1~\cite{xu2018sql} & 34.28  \\
            Graph2Seq2~\cite{xu2018sql} & 38.97   \\
            BART~\cite{lewis-etal-2020-bart}  & 39.27\\
            EzSQL + BART (Ours) & \textbf{40.62}   \\
            \hline
            \hline
            BART~\cite{shi2021learning}&  19.34   \\
            EzSQL + BART (Ours) & \textbf{27.49}   \\
            \hline
        \end{tabular}
        }
    \caption{SQL-to-text results on the WikiSQL (above) and Spider (below).}
    \label{tab:main_results}
\end{table}

\subsubsection{Results and Discussion} Table~\ref{tab:main_results} summarizes the results of our models and baselines. On the WikiSQL dataset, our model outperforms the existing approaches that use different kinds of embedding methods. This shows that the EzSQL IR can bypass the need for any special embedding methods, such as Tree or Graph embeddings for the SQL. Although several SQL-to-text models exist for the simple WikiSQL dataset, few models exist for the smaller and more complex Spider dataset. Recently, approaches use BART as the translator, mainly because generating texts using such a small dataset is not trivial, and taking advantage of pre-trained models can be seen as one of the solutions. The results show that our model performs significantly better than the existing baseline on this dataset. Both models use the same BART-large model, however we use the EzSQL IR as the input instead of the raw SQL. Our approach to bridging the gap between SQL and the NL targets using EzSQL is important, as the pre-trained encoders expect a string of text as input, whereas the existing approach of feeding complex SQLs as a string of text further complicates the task of the encoder. Our model pipeline with EzSQL is much simpler and easier to train since it can efficiently reuse all the parameters of the pre-trained Seq2seq model and does not introduce any new model parameters.

\begin{figure}[h!]
	\centering
	\includegraphics[width=0.8\textwidth]{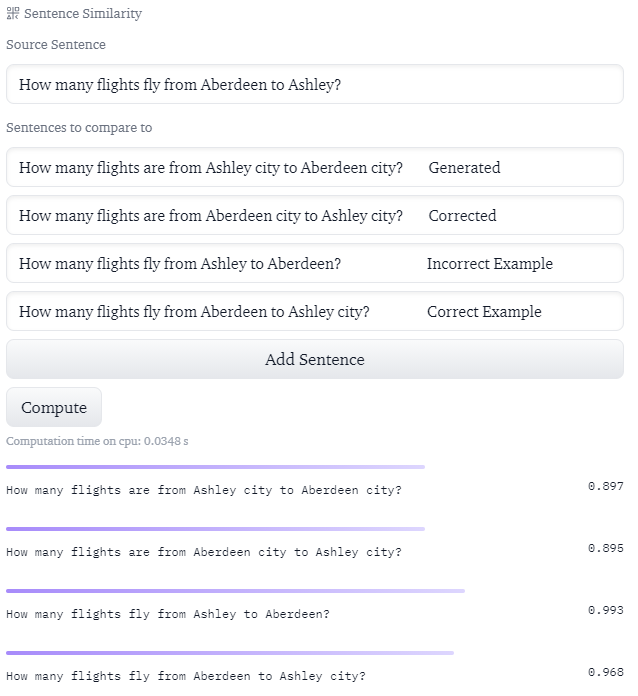}
	\caption{Model error analysis with Sentence BERT.}
	\label{fig:ErrorAnalysis}
\end{figure}

\subsubsection{Model Error Analysis} We analyze the probable causes for error in text generation with BERT sentence similarity score using Sentence BERT. We compute the sentence similarity scores between the gold sentence and the generated sentence. For comparison, we also compute the similarity between the gold and other manually corrected, as well as incorrect, narrations of the SQL. The comparison of the scores is shown in Figure~\ref{fig:ErrorAnalysis}. The erroneous narration and the corrected version have almost the same similarity score. This indicates that our model will also not have enough information to correct the error based on the target only. In the last two examples, a few alterations in the narration can change or flip the scores, confusing the model and leading to the incorrect version as output. Here, the input query is ``SELECT count(*) FROM FLIGHTS AS T1 JOIN AIRPORTS AS T2 ON T1.DestAirport  =  T2.AirportCode JOIN AIRPORTS AS T3 ON T1.SourceAirport  =  T3.AirportCode WHERE T2.City  =  ``Ashley'' AND T3.City  =  ``Aberdeen'''', where the order of information is ``to-from''. However, we expect the language models to have an overwhelming majority of examples in the ``from-to' order. Hence, the generated text is in the incorrect order, and the target loss cannot correct the error, as we have seen in the analysis. The only way to correct the generation is by reordering the information according to the bias of language models in the intermediate representation.

\subsection{Text-to-SQL}

\begin{table*}[h]
\centering
    \resizebox{1\columnwidth}{!}{
        \begin{tabular}{l|cc}
            \hline
            Model & \textsc{Exact Match}  \\
            \hline
            RAT-SQL$^\spadesuit$ \cite{wang-etal-2020-rat} & 69.7  \\
            RYANSQL$^\spadesuit$ \cite{choi2020ryansql} & 70.6 \\
            IRNet$^\diamondsuit$ \cite{guo-etal-2019-towards} & 61.9  \\
            GAZP \cite{zhong-etal-2020-grounded} & 59.1    \\
            BRIDGE$^\spadesuit$  \cite{lin-etal-2020-bridging} & 70.0   \\
            \hline
            RAT-SQL $^\diamondsuit$ \cite{zhang2021data} &  65.4   \\
            RAT-SQL $^\diamondsuit$  with Augmented Data \cite{zhang2021data}& 68.2   \\
            \hline
            RAT-SQL $^\clubsuit$ \cite{wang2021learning}&  70.4   \\
            RAT-SQL $^\clubsuit$ + Pre-Train \cite{wang2021learning}& 71.8   \\
            \hline
            RAT-SQL $^\diamondsuit$ + Pre-Train with our model data & \textbf{69.6}   \\ 
            RAT-SQL $^\clubsuit$ + Pre-Train with our model data & \textbf{73.3}   \\ 
            \hline
        \end{tabular}
        }
    \caption{Set match accuracy on Spider. $^\spadesuit$ stands for models with BERT-large, $^\diamondsuit$ for BERT-base, $^\clubsuit$ for Electra-base.}
    \label{tab:main_results_spider}
\end{table*}

We further experimented on the effectiveness of our model in pretraining data generation for Text-to-SQL parsers. Following~\cite{wang2021learning}, we use the database-specific probabilistic context-free grammar (PCFG) model to sample new SQL queries from the Spider dataset. In this method, ASDL~\cite{wang1997zephyr}  grammar of SQL is represented by a set of context-free grammar (CFG) rules. The grammar is pre-defined based on expert knowledge, and it has been successfully applied in previous work~\cite{yin-neubig-2018-tranx,wang-etal-2020-rat} on Text-to-SQL parsing. A separate PCFG is trained on each source database based on the assumption that each database has a different distribution of SQL queries. 
Using our SQL-to-text model trained on the Spider dataset, we generate the questions from the newly sampled SQLs.

The size of the synthesized data is kept proportional to the size of the original data and the ratio is the same as that in the baseline. Using this synthesized data, we pre-train a RAT-SQL~\cite{wang-etal-2020-rat} semantic parser with the same pre-training configurations and parameters used by~\cite{wang2021learning}. Finally, the parser is fine-tuned on the training set of the Spider dataset. We repeat the experiment using BERT-base for comparison to the recent data augmented training baseline of \cite{zhang2021data}.

\subsubsection{Results and Discussion} The evaluation results of the parser are shown in Table~\ref{tab:main_results_spider}. It can be seen that the RAT-SQL parser pre-trained using the synthesized data with the help of our SQL-to-text model improves the performance of the RAT-SQL baseline. Moreover, the parser outperforms the pre-trained RAT-SQL of \cite{wang2021learning}, although the same method of data synthesis is used in both models. 

The results in Table~\ref{tab:main_results_spider} identify whether the EzSQL IR method is necessary and if it helps to improve the final Text-to-SQL parsing performance. Although in the SQL-to-text experiments we have seen significant improvement in performance, the goal of the SQL-to-text task, going forward, is to assist in data augmentation and data generation for semantic parsers. The improvement of performance over the existing RAT-SQL pre-train model by introducing EzSQL IR as the only additional factor shows the effectiveness of our enhanced SQL-to-text model in data generation. 

\subsection{Ablation Study}

\begin{figure}[h!]
	\centering
\includegraphics[width=0.8\textwidth]{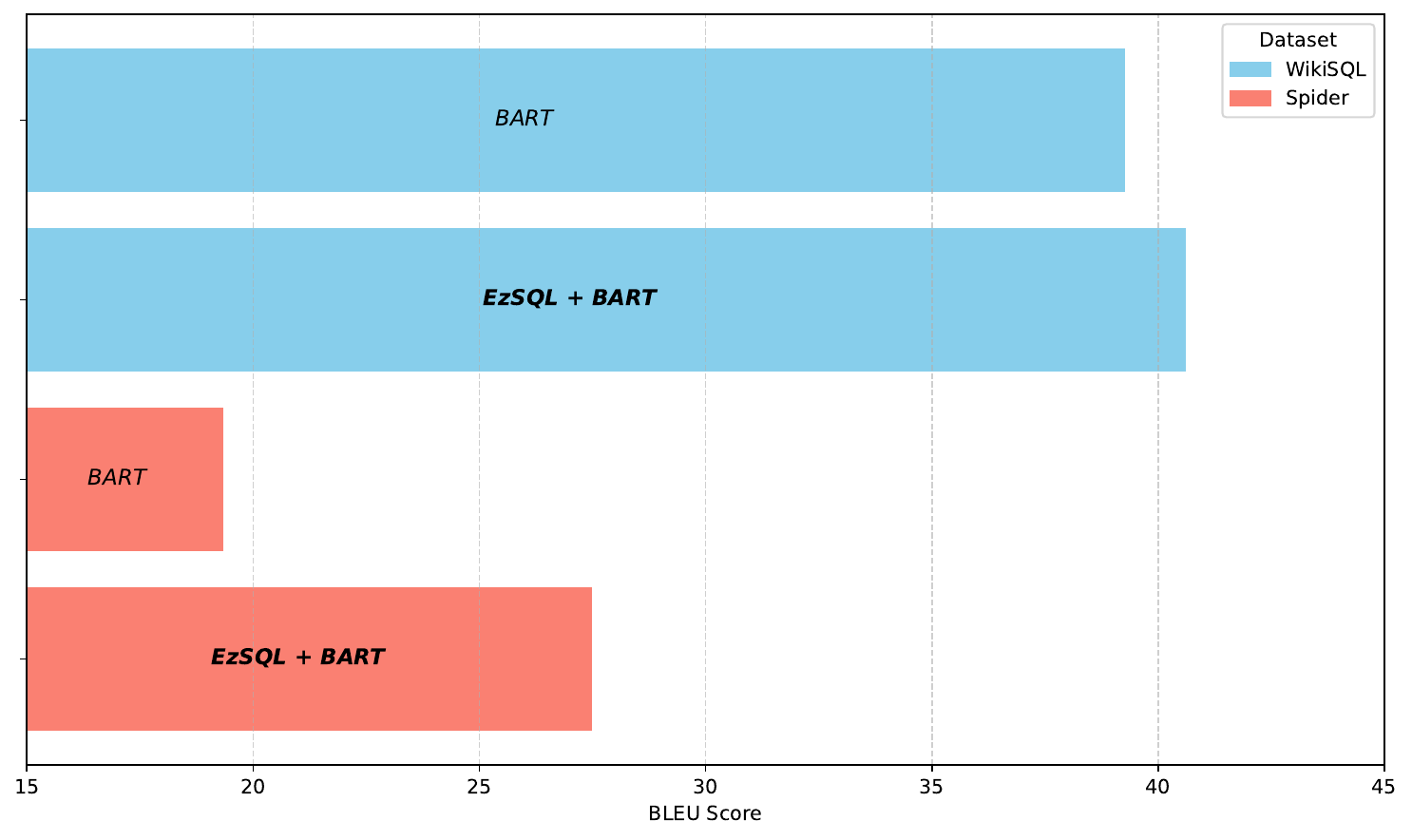}
	\caption{Ablation study for SQL-to-Text translation task.}
	\label{fig:Ablation1}
\end{figure}

\begin{figure}[h!]
	\centering
\includegraphics[width=0.8\textwidth]{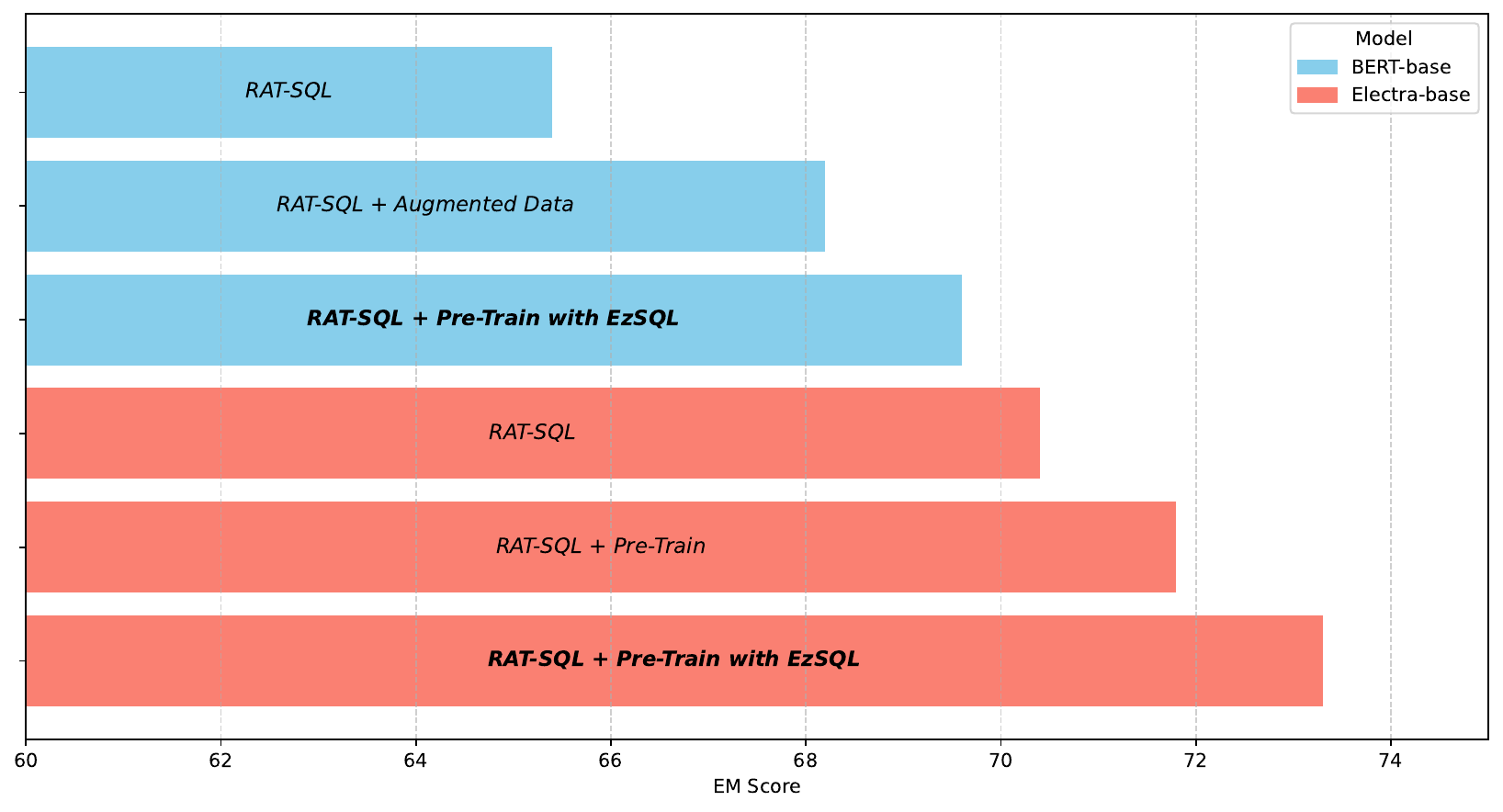}
	\caption{Ablation study for Text-to-SQL translation task.}
	\label{fig:Ablation2}
\end{figure}

We perform ablation studies for both the SQL-to-Text and Text-to-SQL tasks to understand the effect of our proposed method on the performance of the translations. Figure~\ref{fig:Ablation1} shows the ablation study for SQL-to-text translation. We perform this study by taking the EzSQL model with BART and then analyzing the performance without the proposed EzSQL model. This is done for two standard datasets, viz. WikiSQL and Spider. BART was used in this experiment to make a fair comparison of the performance with existing models such as~\cite{lewis-etal-2020-bart,shi2021learning} that use the same BART model.

Figure~\ref{fig:Ablation2} shows the ablation study for Text-to-SQL translation. Similar to the previous study, we take the well-known RAT-SQL model pre-trained with our EzSQL IR and compare the performance without our method on the Spider dataset. In this experiment, we use the BERT and Electra models to match the settings of existing baseline and state-of-the-art models such as \cite{wang2021learning} and \cite{zhang2021data}.

As illustrated in Figures~\ref{fig:Ablation1} and \ref{fig:Ablation2}, we see that the proposed model has a significant contribution to improving the performance of the SQL-to-text and Text-to-SQL translation.

\section{Conclusion and Future Work}
We proposed a new SQL intermediate representation aimed at improving the SQL-to-text task that is simple yet effective for enhancing translation performance using pre-trained language models. The proposed SQL-to-text model was able to outperform the baselines on both WikiSQL and Spider datasets. Moreover, we successfully show that the proposed model can be used to synthesize pre-training data to further enhance the performance of Text-to-SQL semantic parsers. The proposed EzSQL method also illustrates several advantages. EzSQL produces an intermediate representation (IR) of any SQL, and this IR can be used to train any existing seq2seq model and enhance its performance. The method is simple and has little overhead in the generation of the IRs. The flexibility and modularity introduced by the proposed method can enhance Text-to-SQL tasks as well by generating augmented data as shown in the experiments.
However, EzSQL introduces an extra step in the end-to-end training of a seq2seq model, where we need to introduce the IR to the model being trained. Although the EzSQL model works for most of the popular Text-to-SQL datasets, some custom databases may need some additional settings in the proposed EzSQL model to produce the best results. In the future, we intend to add natural keywords to EzSQL, which relay the semantic sense more clearly, that current LLMs tend to struggle with. For example, when given the SQL -`` SELECT T1.AirportCode FROM AIRPORTS AS T1 JOIN 
FLIGHTS AS T2 ON T1.AirportCode  =  T2.DestAirport 
OR T1.AirportCode  =  T2.SourceAirport GROUP BY 
T1.AirportCode ORDER BY count(*) LIMIT 1 '' to GPT-4, it gives the output as - ``Which airport is involved in the highest number of flights, either as a source or destination airport?''. This is wrong, as the order by is ascending by default, and the output should be the fewest number of flights. Thus, these errors would be avoided if a natural language keyword (like ascending in this case) is included in an intermediate form, like EzSQL.


\bibliography{anthology,custom}
\bibliographystyle{model5-names}
\biboptions{authoryear}

\end{document}